\documentclass[lettersize,journal,twoside]{IEEEtran}
\usepackage{amsmath,amsfonts,amssymb}
\usepackage[ruled,norelsize]{algorithm2e}
\makeatletter
\newcommand{\removelatexerror}{\let\@latex@error\@gobble}
\makeatother
\SetKwComment{Comment}{/* }{ */}
\usepackage{array}
\usepackage{textcomp}
\usepackage{stfloats}
\usepackage{url}
\usepackage{verbatim}
\usepackage{graphicx}
\usepackage{cite}

\usepackage{dsfont}

\usepackage[caption=false,listofformat=subsimple,labelformat=simple]{subfig}

\renewcommand{\thesubsubsection}{\Roman{section}-\Alph{subsection} \arabic{subsubsection})}

\newcounter{inlinesubsubsection}
\newcommand{\inlinesubsubsection}[1]{%
    \refstepcounter{inlinesubsubsection}%
    \textit{\theinlinesubsubsection) #1:}%
}

\usepackage{multirow}
\usepackage{multicol}

\usepackage{xcolor}

\usepackage{hyperref}

\usepackage{titlesec}
\usepackage{indentfirst}
\titlespacing*{\section}{0pt}{1.4em}{0.7em}
\titlespacing*{\subsection}{0pt}{0.4em}{0.2em}
\titlespacing*{\subsubsection}{0pt}{0.2em}{0.1em}

\begin{document}
\title{Point Cloud Structural Similarity-based\\Underwater Sonar Loop Detection}

\author{Donghwi Jung$^{1}$, \emph{Graduate Student Member, IEEE}, Andres Pulido$^{2}$, Jane Shin$^{2}$, \emph{Member, IEEE},\\and Seong-Woo Kim$^{1}$, \emph{Member, IEEE}%
\thanks{Received 16 September 2024; accepted 20 February 2025. Date of publication 3 March 2025; date of current version 13 March 2025. This article was recommended for publication by Associate Editor Yue Wang and Editor Sven Behnke upon evaluation of the reviewers’ comments. This work was supported by the Korea Institute for Advancement of Technology through the Korea Government(MOTIE) under Grant P0017304, in part by the Human Resource Development Program for Industrial Innovation, and in part by the Korean Ministry of Land, Infrastructure and Transport(MOLIT) as the Innovative Talent Education Program for Smart City. The Institute of Engineering Research at Seoul National University provided research facilities for this work. \emph{(Corresponding author: Seong-Woo Kim.)}} %Use only for final RAL version
\thanks{$^{1}$Donghwi Jung and Seong-Woo kim are with Seoul National University, Seoul, South Korea. {\tt\footnotesize \{donghwijung,snwoo\}@snu.ac.kr}}%
\thanks{$^{2} $Andres Pulido and Jane Shin are with University of Florida, Florida, United States. {\tt\footnotesize \{andrespulido,jane.shin\}@ufl.edu}}%
\thanks{Our code is available at https://github.com/donghwijung/point\_cloud\_structu\\ral\_similarity\_based\_underwater\_sonar\_loop\_detection.}
\thanks{Digital Object Identifier 10.1109/LRA.2025.3547304}
}
 
% The paper headers
% \markboth{IEEE Robotics and Automation Letters. Preprint Version. Accepted February, 2025}
\markboth{IEEE ROBOTICS AND AUTOMATION LETTERS, VOL. 10, NO. 4, APRIL 2025}
{Jung \MakeLowercase{\textit{et al.}}: Point Cloud Structural Similarity-based Underwater Sonar Loop Detection}
% \markboth{Left-Aligned Header (Odd Pages)}{Right-Aligned Header (Even Pages)}

\maketitle

\begin{abstract}
In this letter, we propose a point cloud structural similarity-based loop detection method for underwater Simultaneous Localization and Mapping using sonar sensors. Existing sonar-based loop detection approaches often rely on 2D projection and keypoint extraction, which can lead to data loss and poor performance in feature-scarce environments. Additionally, methods based on neural networks or Bag-of-Words require extensive preprocessing, such as model training or vocabulary creation, reducing adaptability to new environments. To address these challenges, our method directly utilizes 3D sonar point clouds without projection and computes point-wise structural feature maps based on geometry, normals, and curvature. By leveraging rotation-invariant similarity comparisons, the proposed approach eliminates the need for keypoint detection and ensures robust loop detection across diverse underwater terrains. We validate our method using two real-world datasets: the Antarctica dataset obtained from deep underwater and the Seaward dataset collected from rivers and lakes. Experimental results show that our method achieves the highest loop detection performance compared to existing keypointbased and learning-based approaches while requiring no additional training or preprocessing.
\end{abstract}
\begin{IEEEkeywords}
Bathymetry, loop detection, point cloud, sonar, underwater.
\end{IEEEkeywords}
\section{Introduction}
\IEEEPARstart{A}{ccurate} 3D mapping of the environment is essential for the autonomous navigation\cite{kim2017autonomous}, typically achieved using g point cloud-based simultaneous localization and mapping (SLAM)~\cite{zhang2014loam,palomer2016multibeam,shan2018lego,torroba2019towards,cheng2022underwater,zhang2024ttt}. However, as the travel distance increases, pose estimation errors accumulate during the SLAM process. Loop closure related techniques \cite{kim2018scan,jiang2023contour,yuan2023std,hammond2015automated,tan2023data,zhang2024shape} are employed to mitigate these errors by introducing constraints between nodes in the pose graph when the current position matches a previously visited location, thereby optimizing the pose graph and enabling the generation of accurate 3D maps. Existing loop detection methods, described in Table \ref{table:related_works}, include LiDAR-based approaches \cite{kim2018scan,jiang2023contour,yuan2023std} that utilize dense, multi-channel LiDAR point clouds with a 360-degree Field of View (FoV). These point clouds, typically comprising tens of thousands of points per sequence, offer sufficient detail to enable reliable loop detection without requiring additional data accumulation. Furthermore, they provide distinguishable features, such as buildings and vehicles, which are not only useful for loop detection but also applicable to various autonomous driving algorithms\cite{shin2017real,jung2021uncertainty,woo2024no} in ground environments. For autonomous navigation in underwater environments, however, the use of LiDAR faces significant challenges, due to issues such as laser scattering \cite{guo2021errors}.\\
\indent As a result, sonar is predominantly employed instead of LiDAR for generating point clouds\cite{cho2017auv} in underwater autonomous navigation applications. Sonar-generated point clouds are sparse\cite{sung2020underwater,pulido2022time}, typically consisting of fewer than 1,000 points per sequence, and lack distinct features, such as keypoints\cite{serafin2016fast}. Consequently, LiDAR-based loop detection methods are generally ineffective in underwater applications. Therefore, sonar point cloud-based loop detection algorithms specifically tailored for underwater environments have been proposed \cite{hammond2015automated,tan2023data,zhang2024shape}. Sonar-based approaches include methods that project point clouds into 2D images for processing \cite{hammond2015automated} and those that directly use point clouds as input \cite{tan2023data,zhang2024shape}. While projection-based methods may experience information loss due to point overlap during the transformation process \cite{yang2020predicting}, direct point cloud-based methods avoid this issue. However, these direct methods often require extensive preprocessing, including training environment-specific neural network models or creating new vocabularies tailored to the loop detection task.\\
\indent This letter proposes a method for loop detection based on the structural similarity of 3D point clouds derived from Multibeam Echosounder (MBES) data. The downward viewing direction of the MBES was chosen to address the scarcity of horizontal features in deep-sea environments \cite{lohmann1996orientation}, as this orientation captures relatively more features. To account for the minimal overlap between point clouds captured at the same location from different directions due to the linear distribution of single-sequence sonar data, consecutive point clouds are accumulated before and after the sequence of interest. Square cropping in the $x,y$ directions is applied to extract relevant data, demonstrating improved overlap and similarity comparisons compared to cylindrical cropping \cite{tan2023data}.
% To overcome limitations of learning-based methods reliant on training data,
Moreover, our proposed approach computes point cloud similarity using feature maps based on structural properties derived from spatial relationships among neighboring points, as outlined in PointSSIM \cite{alexiou2020towards}.
These feature maps are rotation-invariant, enabling effective loop detection even when point clouds are captured from varying orientations at the same location. Loop pairs with high similarity exceeding a predefined threshold are identified, eliminating the need for models or predefined vocabularies. This approach demonstrates broad applicability across diverse environments, including deep seas, rivers, and lakes. Additionally, by leveraging point-wise structural feature maps, the method shows robust performance in environments with simple surface slopes, outperforming traditional keypoint-based methods.\\
\begin{table*}[!t]\label{table:related_works}
\caption{Comparison With the Representative Previous Works in Loop Detection Using Point Clouds.}
\centering
\def\arraystretch{1.5}%
\setlength{\tabcolsep}{7.5pt}
\begin{tabular}{|c|c|c|c|c|c|c|c|}
\hline
& Sensor& Description & Bathymetry& w/o 2D Projection & w/o Keypoints & w/o Preprocessing\\
\hline
Kim \emph{et al.}\cite{kim2018scan} &LiDAR&Polar images matching& &\checkmark&\checkmark&\checkmark\\
\hline
Jiang \emph{et al.}\cite{jiang2023contour}&LiDAR&BEV topological graph& &\checkmark&\checkmark&\checkmark\\
\hline
Yuan \emph{et al.}\cite{yuan2023std}&LiDAR&3D Triangle descriptor& &&&\checkmark\\
\hline
Hammond \emph{et al.}\cite{hammond2015automated} &Sonar& Image keypoints matching &\checkmark&&& \checkmark\\
\hline
Tan \emph{et al.}\cite{tan2023data} &Sonar& Point cloud keypoints matching & \checkmark&\checkmark &&\\
\hline
Zhang \emph{et al.}\cite{zhang2024shape} &Sonar& Gradient features and BoW & \checkmark&& \checkmark&\\
\hline
This work &Sonar& Point cloud structural similarity & \checkmark&\checkmark& \checkmark& \checkmark\\
\hline
\end{tabular}
\end{table*}
\indent The contributions of our letter are as follows:
\begin{itemize}
    \item We leverage point-wise structural similarity to achieve high-performance loop detection, even in sparse point clouds with few keypoints in bathymetric environments.
    \item We use rotation-invariant feature maps to detect loops, enabling the identification of the same location revisited at different orientations.
    \item We validate the performance of the proposed method using datasets collected from oceans, lakes, and rivers, confirming its applicability across diverse marine settings without additional preprocessing.
\end{itemize}
\section{Related Works}
Existing LiDAR-based point cloud loop detection methods include \cite{kim2018scan,jiang2023contour,yuan2023std}. First, Kim \emph{et al.}\cite{kim2018scan} project the point cloud into a polar image and identifies the most similar loop pair through column-wise matching between images. Jiang \emph{et al.}\cite{jiang2023contour}, on the other hand, convert the raw point cloud into multiple hierarchical Bird's-Eye View (BEV) images based on height. Each BEV image is clustered into elliptical shapes, and these clusters are transformed into graphs with nodes representing clusters. Loop pairs are then identified by comparing these graphs. Lastly, Yuan \emph{et al.}\cite{yuan2023std} utilize three neighboring points to form a triangle, which is then designated as a keypoint. Descriptors such as the triangle's vertices and normal vectors are calculated, and loop pairs are detected by comparing these descriptors. This triangle-based descriptor is invariant to rotation and translation, and the uniqueness of triangles formed by three points makes this approach well-suited for loop detection. However, these LiDAR-based methods face limitations in underwater environments due to laser scattering, making the use of LiDAR impractical\cite{guo2021errors}. Moreover, as these approaches are designed primarily for ground vehicles using 360-degree horizontal LiDAR, they are unsuitable for bathymetric scenarios where point clouds are acquired as flat planes.\\
\indent In underwater scenarios, among prior loop detection approaches using sonar point clouds\cite{hammond2015automated,tan2023data,zhang2024shape}, Hammond \emph{et al.}\cite{hammond2015automated} projected 3D sonar point clouds into 2D and applied image-based feature matching techniques, such as SIFT \cite{lowe2004distinctive}, for loop detection. While this approach performed well in environments with significant depth variations, such as uneven seabeds, it suffered from information loss during projection and was less effective in flatter environments like rivers or lakes, where fewer keypoints were detected. Tan \emph{et al.}\cite{tan2023data} proposed a method that directly uses point clouds as input without projection. This approach aggregates multiple consecutive point clouds through dead reckoning and applies cylindrical cropping to ensure consistent overlap irrespective of vehicle orientation. Although this method simplifies loop detection and reduces reliance on orientation, the overlap area diminishes with increased loop distance, reducing similarity accuracy. Furthermore, it heavily depends on a neural network trained specifically for each environment, requiring additional preprocessing and parameter differentiation.\\
\begin{figure*}[t]
    \centering
    \framebox{\parbox{0.985\textwidth}{\includegraphics[width=0.985\textwidth]{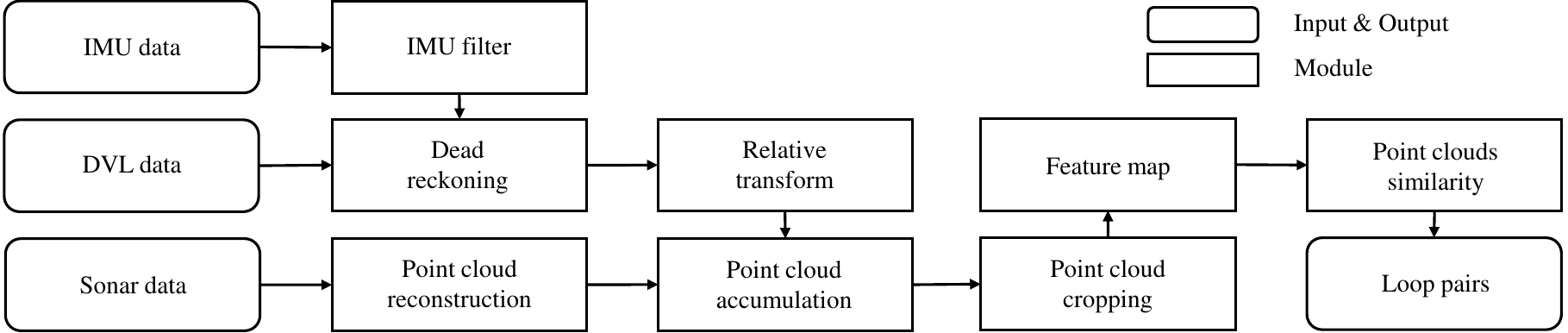}}}
    \caption{Process of our proposed method. The inputs are sensor data from sonar, DVL, and IMU, and the output is the detected loop pairs.}
    \label{fig:process}
\end{figure*}
\indent Zhang \emph{et al.}\cite{zhang2024shape} introduced Shape BoW, a modified Bag-of-Words (BoW) approach, which uses the standard deviation of depth differences between adjacent points for descriptor training and later converts point clouds into images for loop detection. This method avoids keypoint extraction, achieving relatively high accuracy; however, its performance depends on a pre-defined vocabulary created by clustering descriptors into words. Changes in the environmental context alter the descriptor distribution, necessitating the creation of new vocabularies, which limits its adaptability to novel settings.\\
\indent To address these issues, we propose a new method designed to minimize data loss and improve adaptability across diverse environments. Our approach converts the structural features of point clouds into rotation-invariant feature maps and calculates similarity between these maps, enabling accurate loop detection without reliance on 2D projection or keypoint extraction. Experimental results demonstrate that this method outperforms existing techniques while offering broader applicability and reduces dependence on environment-specific preprocessing.
\section{Methods}
Our proposed architecture is depicted in Fig.~\ref{fig:process} and the details of each part are described as follows:
\subsection{Dead Reckoning}\label{methods:dead_reckoning}
First, we perform a dead reckoning computation to estimate the ego-vehicle’s pose changes using linear acceleration, angular velocity, and geomagnetic data collected from the IMU, along with velocity measurements obtained from the Doppler Velocity Log (DVL). This pose estimation is conducted to accumulate the point clouds acquired by the sonar over consecutive sequences. Further details on this process are provided in the point cloud processing section. During this dead reckoning process, various factors can introduce errors in pose estimation, which may affect the overall algorithm performance. For instance, IMU data is subject to sensor noise and environmental disturbances, leading to accumulated errors over time. In particular, errors in pose estimation derived from the IMU increase the inaccuracy of trajectory estimation as the path length grows. To address these issues, we employ the Madgwick filter \cite{madgwick2011estimation} and the Extended Kalman Filter (EKF) \cite{moore2016generalized} during the pose estimation process to minimize errors.

\subsection{Point Cloud Processing}\label{methods:point_cloud_processing}
\begin{figure*}[t]
    \centering
        \subfloat[Accumulated point clouds.]{
    	   \framebox{\parbox{0.7\textwidth}{\includegraphics[width=0.7\textwidth]{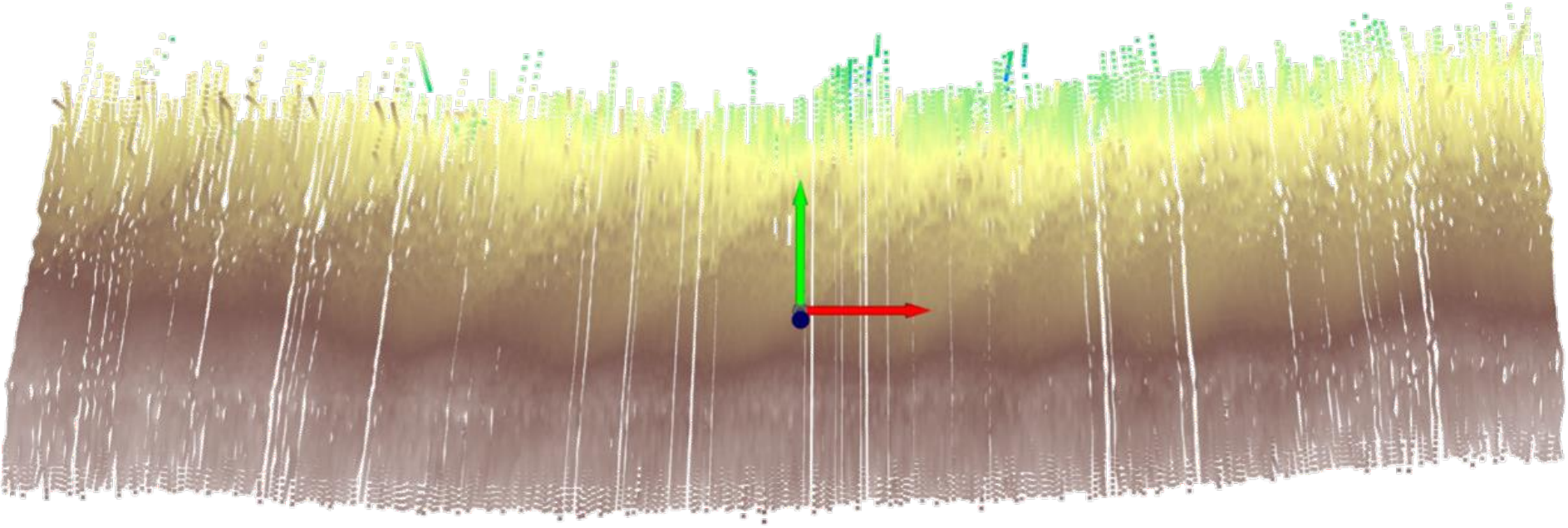}}}\label{fig:accumulated_point_clouds}
        }
        \hfill
        \subfloat[Cropped point cloud.]{
    	   \framebox{\parbox{0.245\textwidth}{\includegraphics[width=0.245\textwidth]{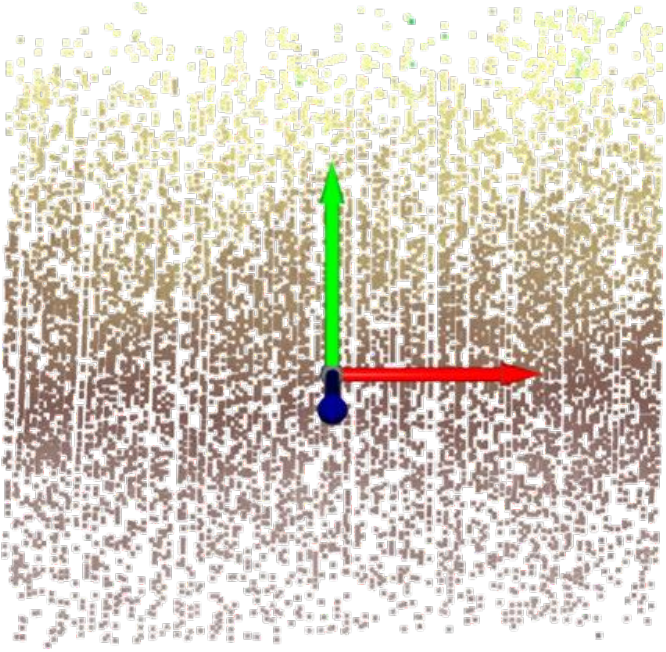}}}\label{fig:cropped_point_cloud}
        }
    \caption{Process of square cropping. (a) Accumulated point clouds based on the estimated poses, (b) Cropped point cloud centered on the pose of vehicle. The colors of the point cloud were assigned arbitrarily based on the z-values of the points to facilitate visualization.}
    \label{fig:point_cloud_processing}
\end{figure*}

In the case of a downward-facing MBES, the beams spread widely in the lateral direction, while they are relatively narrow in the forward and backward directions of the vehicle. Therefore, when reconstructing point clouds from data of a single sequence obtained by MBES, the resulting point cloud, when viewed vertically, resembles single-channel 2D LiDAR data with a limited FoV. In underwater autonomous navigation scenarios, when data is collected downward, there is hardly any overlap when approaching the same location from different directions due to the single-channel nature. Thus, this single-channel point cloud cannot be used to find underwater terrain loops.\\
\indent To address these issues, it is necessary to accumulate consecutive point clouds. In this process, the ego-vehicle’s pose estimated through dead reckoning using DVL and IMU data, as described in Sec.~\ref{methods:dead_reckoning}, serves as the basis for accumulating sequential point clouds. First, relative transformations between the ego-vehicle’s reference pose and other estimated poses are computed. These transformations are then applied to transform and accumulate the sequential point clouds. The accumulated point cloud is subsequently cropped into a rectangular shape within a predefined distance on the $x-y$ plane, using the $z$-axis of the vehicle’s current pose as the reference. The accumulation and square cropping process of the point cloud is illustrated in Fig.~\ref{fig:point_cloud_processing}. This point cloud processing procedure is as follows:
\begin{gather}
    \Delta \mathbf{R}_k=\left(\mathbf{R}_0\right)^{-1}\cdot \mathbf{R}_k,\\
    \Delta \mathbf{t}_k=\left(\mathbf{R}_0\right)^{-1}\cdot \left(\mathbf{t}_k-\mathbf{t}_0\right),\\
    \bar{P}_k = P_k\cdot \Delta \mathbf{R}_k + \Delta \mathbf{t}_k,\\
    \bar{\mathbf{P}} = \bigcup^n_{k=-n}\bar{P}_k,\\
    \mathbf{P}=\{p\;|\;\|p_{x,y}-c_{x,y}\|_2\leq d,\,p\in\bar{\mathbf{P}}\},\;\mathbf{P}\in \mathds{R}^{|\mathbf{P}|\times 3},\label{eqn:square_cropping}
\end{gather}
where $\mathbf{R}$ and $\mathbf{t}$ refer to the orientation and position of the ego vehicle in the global coordinate, respectively. Similarly, $\mathbf{R}_0$ and $\mathbf{t}_0$ signify the orientation and position of the ego vehicle's reference pose. $\Delta \mathbf{R}$ and $\Delta \mathbf{t}$ represent the relative rotation and translation with respect to the reference pose. $P$ denotes the point cloud obtained from the 3D transformation of the raw data from the MBES. $n$ is a predefined value that corresponds to half of the time sequence to accumulate the point cloud. $\bar{P}$ represents the point cloud of a single sequence transformed to the reference pose. $\bar{\mathbf{P}}$ refers to the accumulated point cloud created by transforming the point clouds based on the reference pose using $\Delta \mathbf{R}$ and $\Delta \mathbf{t}$. $\|\cdot\|_2$ indicates the Euclidean distance. $c$ is the reference point of $\bar{\mathbf{P}}$, corresponding to the reference pose of the vehicle. $\left(\cdot\right)_{x,y}$ represents the coordinates $x$ and $y$ of the respective point. $d$ is a predefined value used for square cropping $\bar{\mathbf{P}}$. $\mathbf{P}$ represents the point cloud derived from square cropping, corresponding to a set of all the cropped points. $\bar{\mathbf{P}}$ and $\mathbf{P}$ are visualized in Fig.~\ref{fig:accumulated_point_clouds} and Fig.~\ref{fig:cropped_point_cloud}, respectively.
\begin{figure}[t]
    \centering
    \framebox{\parbox{0.48\textwidth}{\includegraphics[width=0.48\textwidth]{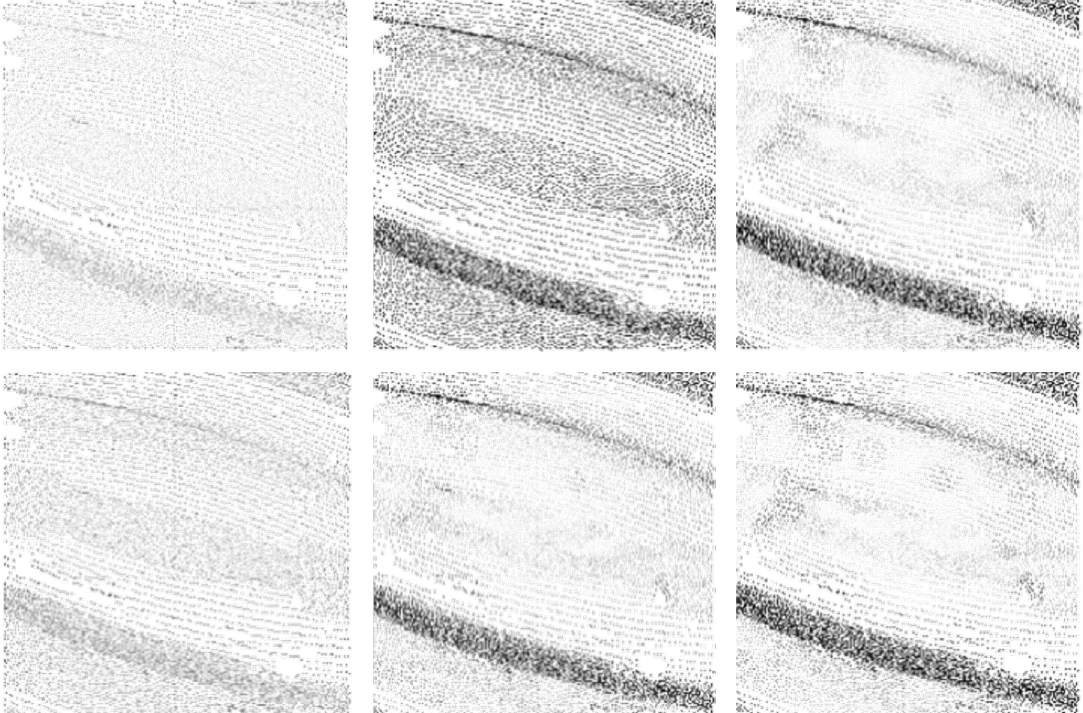}}}
    \caption{Feature maps. Geometry (left), Normal (middle), and Curvature (right). Mean (top) and Variance (bottom). Each feature map, matched to each point in the point cloud, is shown as a 2D grid image for visualization purposes.}
    \label{fig:feature_maps}
\end{figure}
\subsection{Feature Maps}\label{methods:feature_map}
To calculate the similarity between point clouds, we create feature maps using the structural features of the point cloud. Initially, we utilize three types of features: geometry, normals, and curvature. Geometry is represented by the Euclidean distance between each point in the point cloud and its neighboring points. For normals, it involves the angle between the normal vectors of each point and its neighbors. The reason for not directly using the position information of the points and their normal vectors, which are commonly used for geometry and normal features, is that these values are rotation-variant. Because our research also addresses loop detection when passing the same place with different orientations, we use rotation-invariant geometry and normal features. For curvature, it represents the mean curvature for each point. The calculations for the feature maps and the curvature are referenced from \cite{alexiou2020towards} and \cite{meynet2020pcqm}, respectively. The process of creating the quantity, which is the prior stage of the feature map from a square-shaped cropped point cloud is as follows:
\begin{gather}
    \mathbf{G},\mathbf{N},\mathbf{C}=\{G_p\},\{N_p\},\{C_p\}\in \mathds{R}^{|\mathbf{P}|\times M},\quad p\in\mathbf{P},\label{eqn:g_n_c}\\
    G_p=\big\{\|p-p_i\|_2\big\},\quad 0\leq i<M,\\
    N_p=\bigg\{\arccos{\left(\frac{\Vec{n}\cdot\Vec{n}_i}{|\Vec{n}|\cdot|\Vec{n}_i|}\right)}\bigg\},\label{eqn:normal_feature_map}\\
    \rho=\frac{(1+d^2)a+(1+e^2)b-4abc}{(1+e^2+d^2)^\frac{3}{2}},\label{eqn:curvature}\\
    C_p=\{\rho_{p_i}\},
\end{gather}
where $G_p$, $N_p$, and $C_p$ represent the geometry, normal, and curvature quantities at point $p$, respectively. Additionally, $\mathbf{G}$, $\mathbf{N}$, and $\mathbf{C}$ are the sets of $G_p$, $N_p$, and $C_p$ in Eq. \eqref{eqn:g_n_c}. Furthermore, $|\cdot|$, and $|\,\Vec{\cdot}\,|$ denote the cardinality of the set, and the size of the vector, respectively. $p_i$ denotes the $i$-th nearest neighbor point of $p$. $M$ refers to the number of the nearest neighbors. $\Vec{n}$ in Eq. \eqref{eqn:normal_feature_map} represents the normal vector calculated at $p$ by plane fitting and $\Vec{n}_i$ indicates the normal vector at $p_i$. In Eq. \eqref{eqn:curvature}, $a$, $b$, $c$, $d$, and $e$ represent the coefficients of the quadric surface $Q_p(x,y)=ax^2+by^2+cxy+dx+ey+f$, fitted using $p$ and the nearest-neighbor points. This quadric surface fitting is based on the method described in \cite{meynet2020pcqm}.
\begin{figure*}[t]
    \centering
        \subfloat[Point cloud.]{
    	   \framebox{\parbox{0.225\textwidth}{\includegraphics[width=0.225\textwidth]{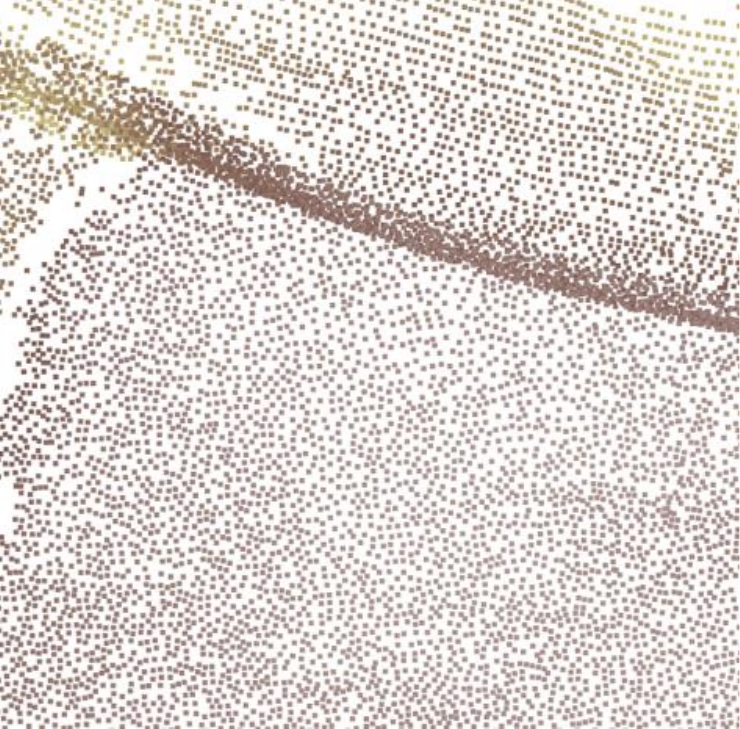}}}
            \label{fig:point_cloud}
        }
        \subfloat[Query.]{
    	   \framebox{\parbox{0.225\textwidth}{\includegraphics[width=0.225\textwidth]{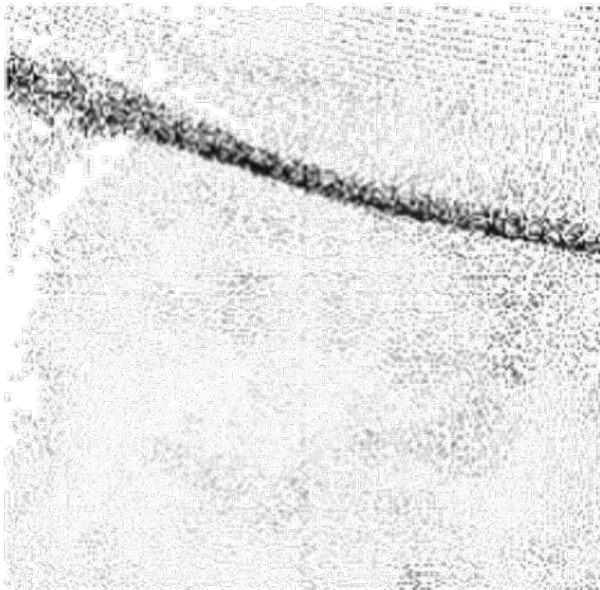}}}
            \label{fig:query_feature_map}
        }
        \subfloat[Positive target.]{
    	   \framebox{\parbox{0.225\textwidth}{\includegraphics[width=0.225\textwidth]{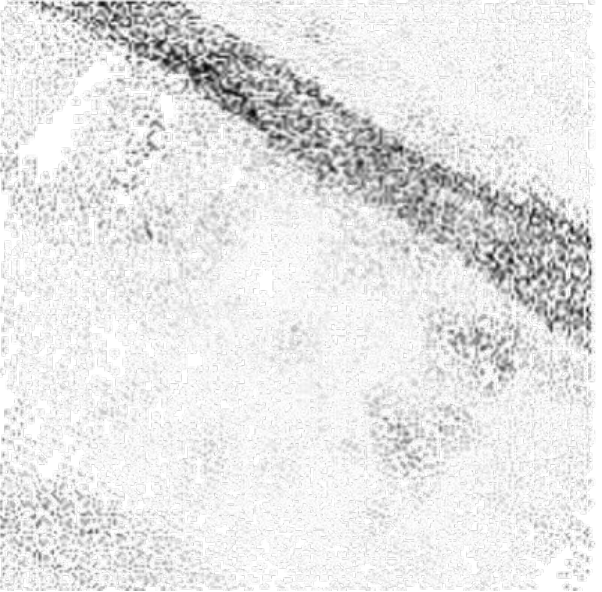}}}
            \label{fig:true_positive_feature_map}
        }
        \subfloat[Negative target.]{
    	   \framebox{\parbox{0.225\textwidth}{\includegraphics[width=0.225\textwidth]{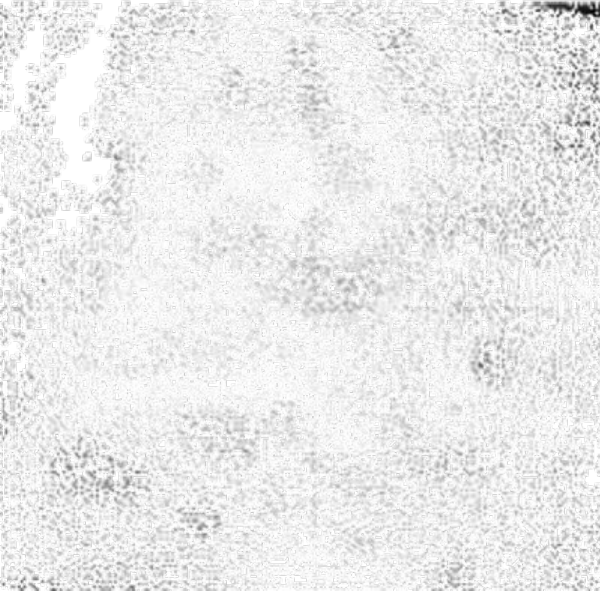}}}
            \label{fig:true_negative_feature_map}
        }
    \caption{Examples of a point cloud and feature maps. From the left, a raw point cloud, a query feature map, the true positive feature map, and the true negative feature map. (a) represents the 3D point cloud, while (b)-(d) depict the feature maps projected onto 2D images.
    }
    \label{fig:data_examples}
\end{figure*}
As shown in Fig.~\ref{fig:feature_maps}, for these calculated quantities, we create feature maps by calculating the mean and variance of the values of each quantity for the neighborhood point by point. By using representative values such as mean and variance, we reduce the dimension of the data used to calculate point cloud similarity, thereby shortening the computation time. The process of calculating these feature maps is as follows:
\begin{gather}
\mathds{F}=\{\mathbf{F}\}\in \mathds{R}^{|\mathbf{P}|\times 2},\\
\mathbf{F}\in\{\mathbf{G}_{\mu},\mathbf{G}_{\sigma^2},\mathbf{N}_{\mu},\mathbf{N}_{\sigma^2},\mathbf{C}_{\mu},\mathbf{C}_{\sigma^2}\},\\
F_{\mu_p}=\frac{1}{|\mathbf{P}|}\sum^{|\mathbf{P}|} F_p,\\
F_{\sigma_p^2}=\frac{1}{|\mathbf{P}|}\sum^{|\mathbf{P}|} \left(F_p-F_{\mu_p}\right)^2,\\
\mathbf{F}_{\mu}=\{F_{\mu_p}\},\;\mathbf{F}_{\sigma^2}=\{F_{\sigma_p^2}\},\quad p\in\mathbf{P},
\end{gather}
where the subscripts $\mu$ and $\sigma^2$ under $\mathbf{G}, \mathbf{N}, \mathbf{C}$ denote the methods used to calculate the feature maps, corresponding to mean and variance, respectively. $\mathbf{F}$ refers to any element of the entire set of feature maps possessed by a single point cloud. $\mathds{F}$ indicates the set of feature maps $\mathbf{F}$. In addition, $F_p$ denotes the feature map $F$ of point $p$. $F_{\mu_p}$ and $F_{\sigma_p^2}$ represent the mean and variance of the feature $F$ at point $p$, respectively, and correspond to elements of $\mathbf{F}_{\mu_p}$ and $\mathbf{F}_{\sigma_p^2}$.
\subsection{Loop Detection}\label{methods:loop_detection}
Our proposed method calculates similarity by comparing the point cloud feature map, computed in Sec.~\ref{methods:feature_map}, with the feature maps of previously stored point clouds. In this case, assuming one of the feature maps is fixed, as the other feature map becomes more similar to the fixed one, the similarity approaches one. Conversely, if the feature maps differ significantly, the similarity converges toward zero. The sum of similarities obtained in this manner is identified and, if it exceeds a predefined threshold, it is considered a loop. This process can be mathematically expressed as follows:
\begin{gather}
    S(p_i,p_j,\mathbf{F})=1-\frac{abs\left(\mathbf{F}(p_j)-\mathbf{F}(p_i)\right)}{\max\{abs\left(\mathbf{F}(p_j)\right),abs\left(\mathbf{F}(p_i)\right)\}+\epsilon},\\
    S(\mathbf{P}_i,\mathbf{P}_j,\mathbf{F})=\frac{1}{|\mathbf{P}_i|}\frac{1}{|\mathbf{P}_j|}\sum^{|\mathbf{P}_i|}\sum^{|\mathbf{P}_j|}S(p_i,p_j,\mathbf{F}),\\
    \Gamma\left(\mathbf{P}_i,\mathbf{P}_j\right) = \sum^{|\mathds{F}|} S(\mathbf{P}_i,\mathbf{P}_j,\mathbf{F}),\\
    f(i,j) = \begin{cases}
    True, & \text{if }\;\Gamma(\mathbf{P}_i,\mathbf{P}_j) > \gamma \\ False, & \text{otherwise} \end{cases},\label{loop_decision}
\end{gather}
where $S(\cdot,\cdot,\cdot)$ is the structural similarity between the first two elements calculated based on the third element. Moreover, $abs$ indicates the absolute value. $\epsilon$ is a constant that is used to prevent the denominator in $S$ from being zero. $\Gamma$ indicates similarity between two point clouds corresponding to the average of similarities of the feature maps. $\gamma$ represents the threshold for each feature map used to determine whether it constitutes a loop. Lastly, $f$ is the indicator function to determine whether two sequences are a loop pair or not.
\section{Experiments}
\subsection{Purposes and Details}
The two research questions we aim to verify in this letter are as follows:
Q1. Does the proposed method demonstrate good loop detection performance in various environments without preprocessing tasks such as model training or creating a vocabulary?
Q2. Does the method perform well even in monotonous underwater environments by utilizing the point-wise information of the 3D point cloud?
To answer the first research question, we compared the proposed method’s loop detection performance against \cite{tan2023data}, which trains a neural network model for each dataset, and \cite{zhang2024shape}, which generates a vocabulary for each dataset. To address the second question, we verified whether the proposed method can successfully detect loops even in environments where few appropriate keypoints are detected. Additionally, for comparative validation, we compared the loop detection performance on the same datasets with a keypoint-based method\cite{hammond2015automated}. Additionally, we compared the proposed approach with existing LiDAR point cloud-based loop detection methods \cite{kim2018scan,jiang2023contour,yuan2023std}, which were validated in ground vehicle environments, to evaluate whether these methods also demonstrate strong performance in underwater loop detection scenarios.\\
\indent In this experiment, we evaluated the performance of the proposed algorithm using the Antarctica dataset, obtained using the Hugin 3000 AUV \cite{tan2023data}, and the Seaward dataset, acquired from banks via boats \cite{krasnosky2022bathymetric}. To compare the performance of the proposed method, we used previous point cloud-based loop detection methods\cite{kim2018scan,jiang2023contour,yuan2023std,hammond2015automated,tan2023data,zhang2024shape} as baselines. Because the codes for \cite{hammond2015automated,zhang2024shape} were not publicly available, we implemented the algorithms based on the descriptions in the respective papers, except for \cite{kim2018scan,jiang2023contour,yuan2023std,tan2023data}. Moreover, to test \cite{tan2023data}, we trained a neural network model using each dataset. Similarly, we created a vocabulary for each dataset and validated the performance of Shape BoW\cite{zhang2024shape}. For the Antarctica dataset, where the depth was greater and the lateral spread was wider, the distance parameter $d$ was set to $100\ m$ based on the criteria in \cite{tan2023data}.
\begin{figure*}[t]
    \centering
        \subfloat[Our method.]{
    	   \framebox{\parbox{0.225\textwidth}{\includegraphics[width=0.225\textwidth]{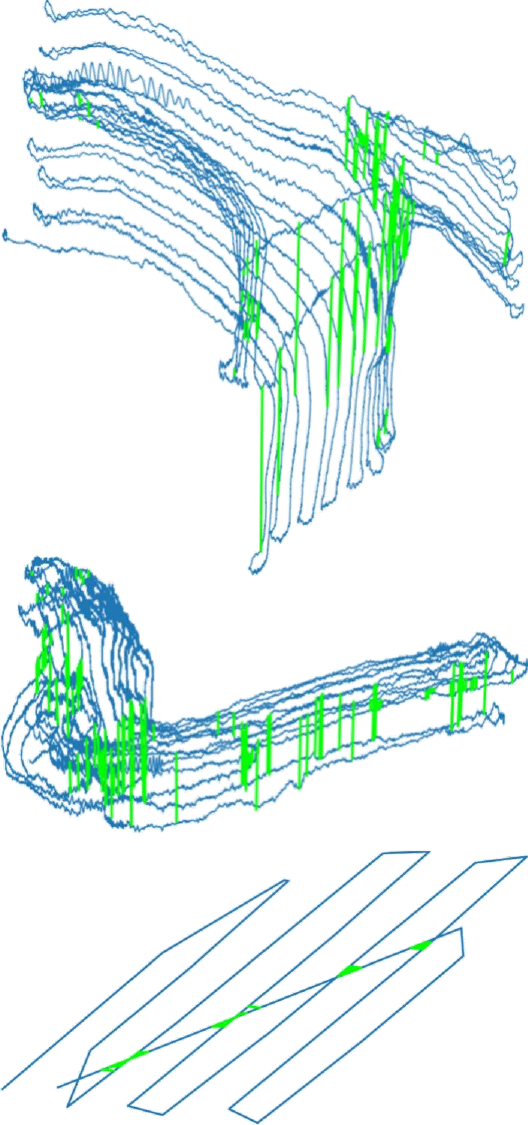}}}
            \label{fig:loop_detection_result_pointssim}
        }
        \subfloat[Hammond \emph{et al.}\cite{hammond2015automated}]{
    	   \framebox{\parbox{0.225\textwidth}{\includegraphics[width=0.225\textwidth]{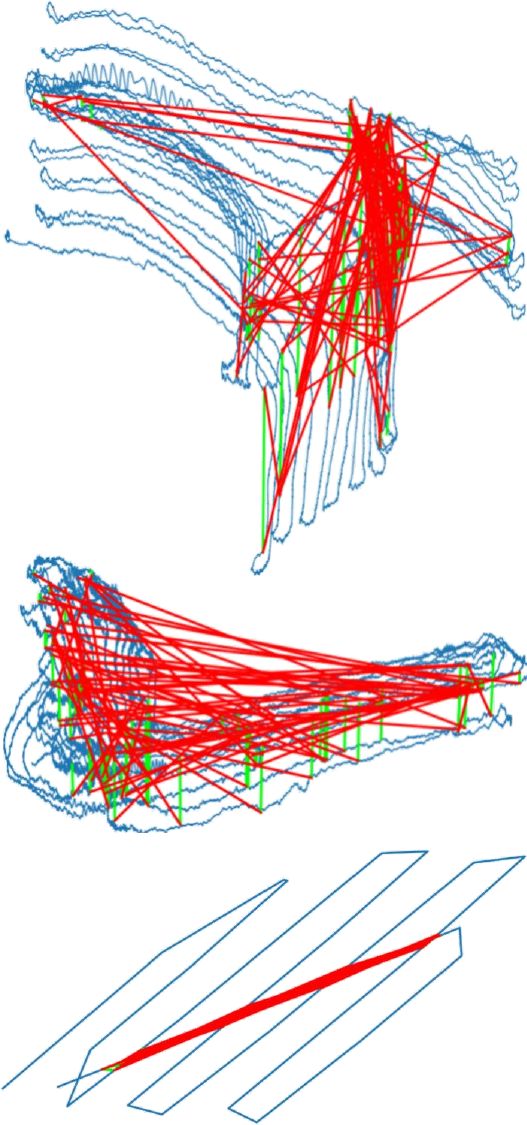}}}
            \label{fig:loop_detection_result_sift}
        }
        \subfloat[Tan \emph{et al.}\cite{tan2023data}]{
    	   \framebox{\parbox{0.225\textwidth}{\includegraphics[width=0.225\textwidth]{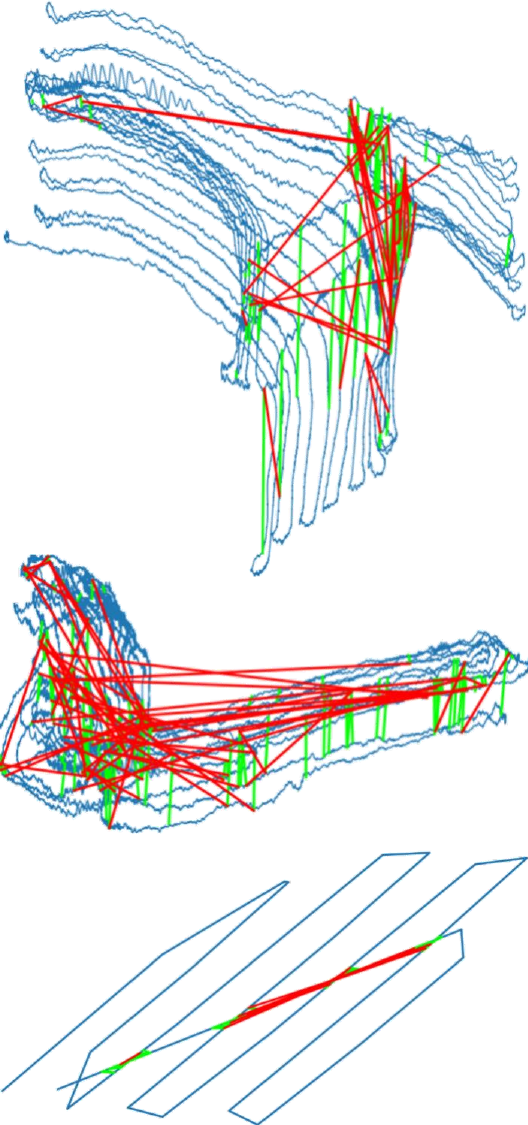}}}
            \label{loop_detection_result_bathy}
        }
        \subfloat[Zhang \emph{et al.}.\cite{zhang2024shape}]{
    	   \framebox{\parbox{0.225\textwidth}{\includegraphics[width=0.225\textwidth]{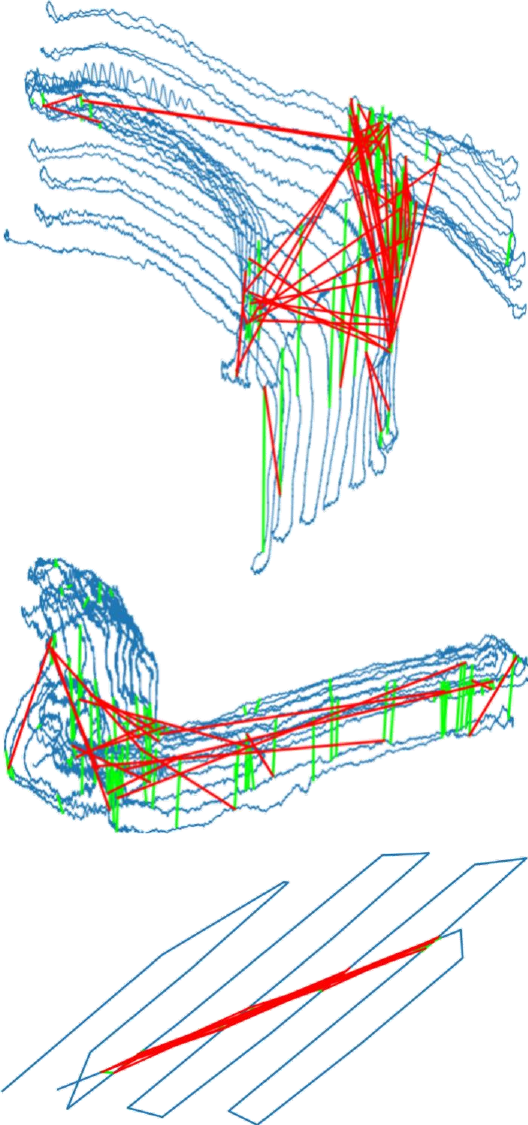}}}
            \label{fig:loop_detection_result_shape}
        }
    \caption{The results of sonar-based loop detection methods. From top to bottom, the trajectories and predicted loop pairs for each method are shown in the datasets of the North River, Wiggles bank, and Antarctica. Green indicates true positive pairs, and red indicates false positive pairs.}
    \label{fig:loop_detection_results}
\end{figure*}
\begin{figure*}[t]
    \centering
        \subfloat[Antarctica.]{
    	   \framebox{\parbox{0.31\textwidth}{\includegraphics[width=0.31\textwidth]{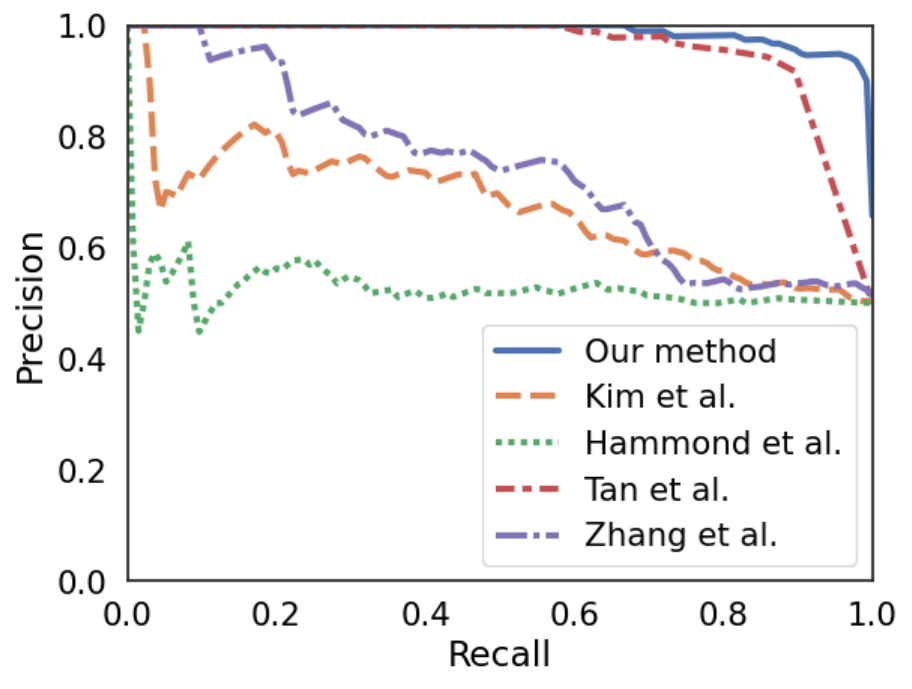}}}
            \label{fig:pr_curve_antarctica}
        }
        \subfloat[Wiggles bank.]{
    	   \framebox{\parbox{0.31\textwidth}{\includegraphics[width=0.31\textwidth]{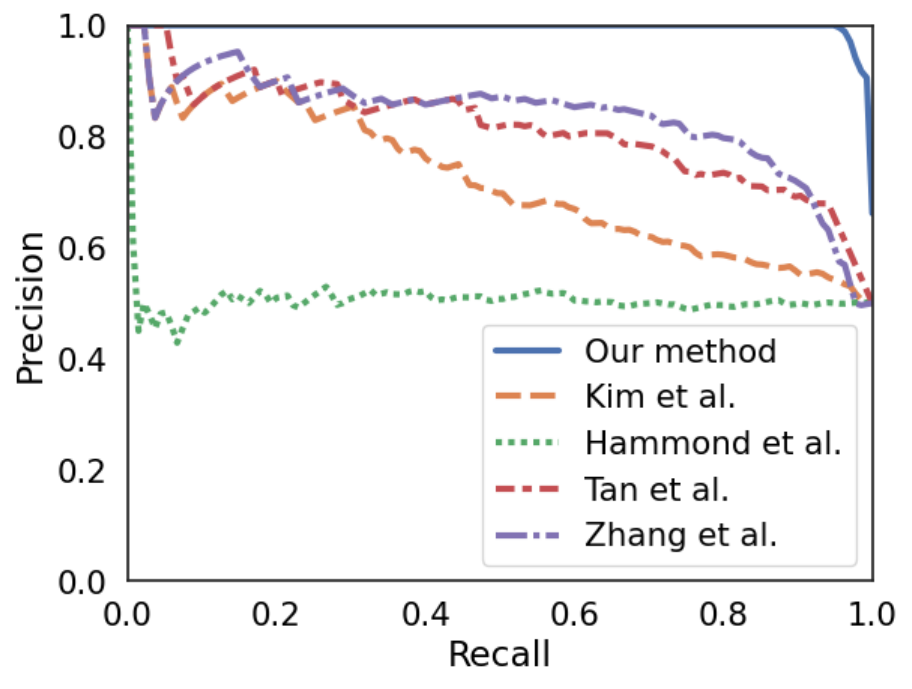}}}
            \label{fig:pr_curve_wiggles_bank}
        }
        \subfloat[North river.]{
    	   \framebox{\parbox{0.31\textwidth}{\includegraphics[width=0.31\textwidth]{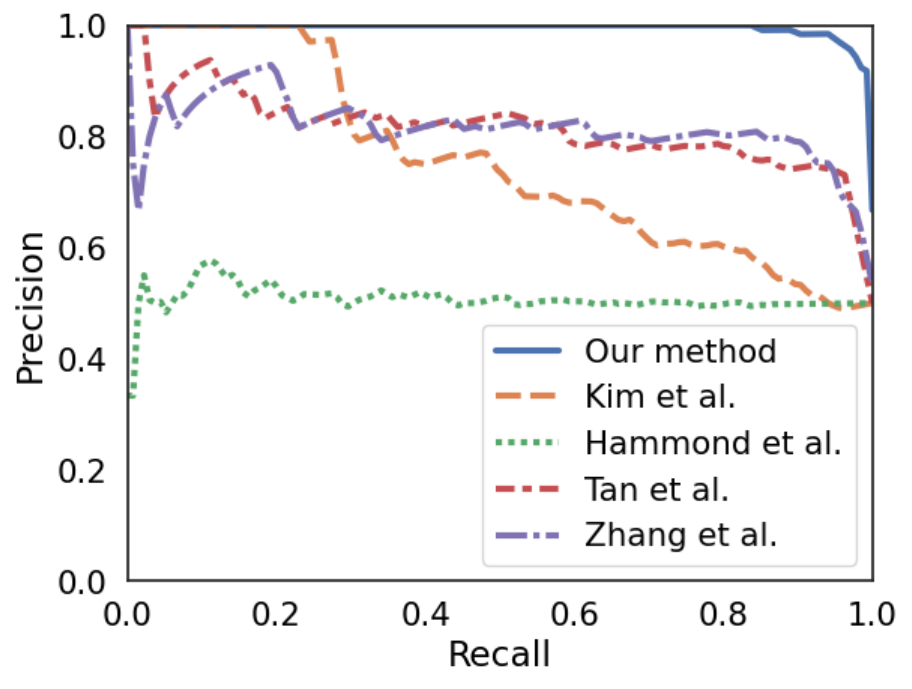}}}
            \label{fig:pr_curve_north_river}
        }
        \hfill
        \vspace{-0.4cm}
        \subfloat[Dutch island.]{
    	   \framebox{\parbox{0.31\textwidth}{\includegraphics[width=0.31\textwidth]{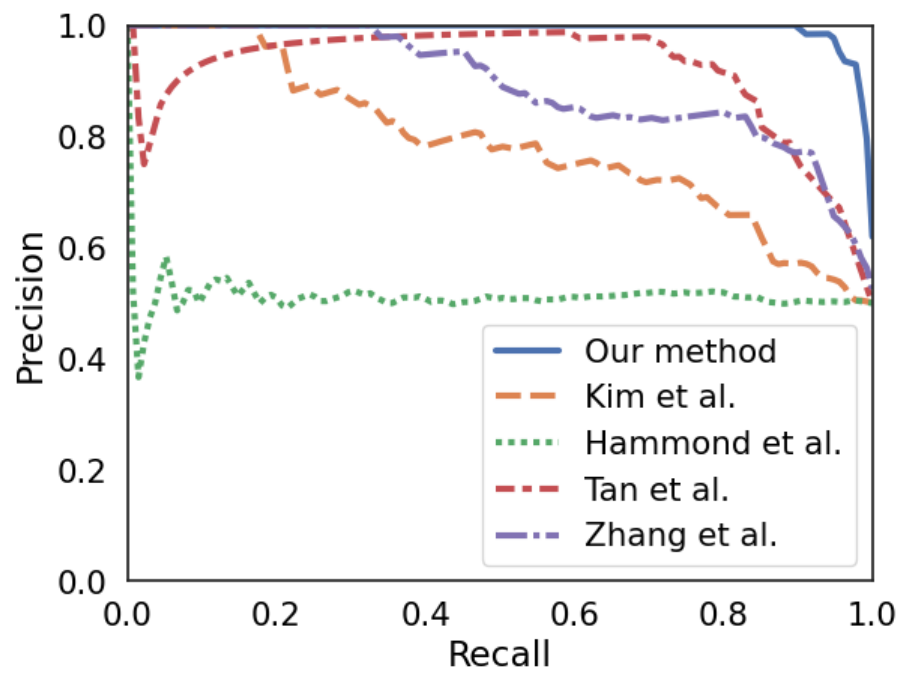}}}
            \label{fig:pr_curve_dutch_island}
        }
        \subfloat[Beach pond.]{
    	   \framebox{\parbox{0.31\textwidth}{\includegraphics[width=0.31\textwidth]{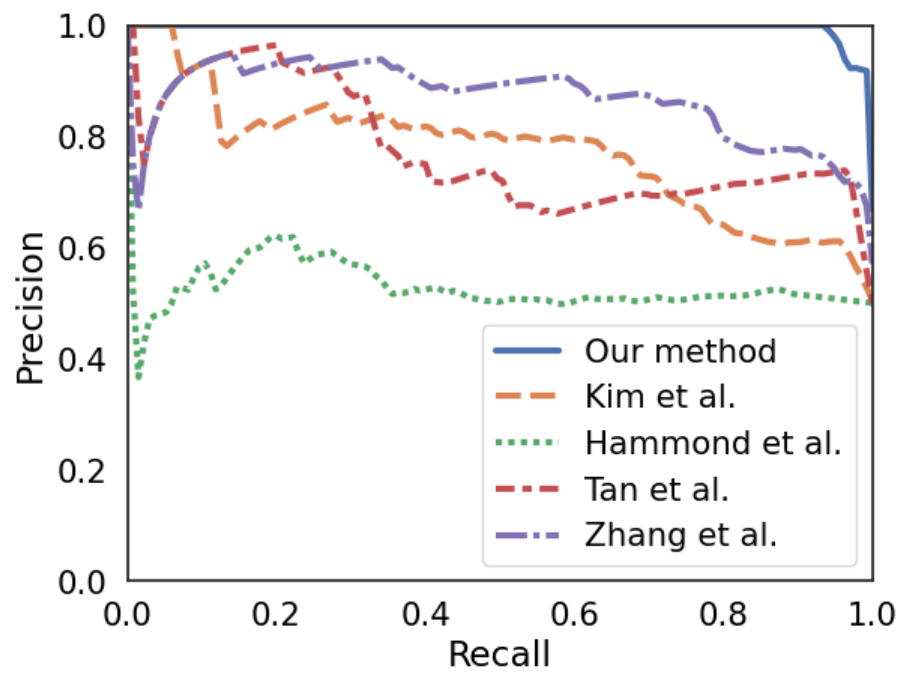}}}
            \label{fig:pr_curve_beach_pond}
        }
        \subfloat[Ablation study.]{
    	   \framebox{\parbox{0.31\textwidth}{\includegraphics[width=0.31\textwidth]{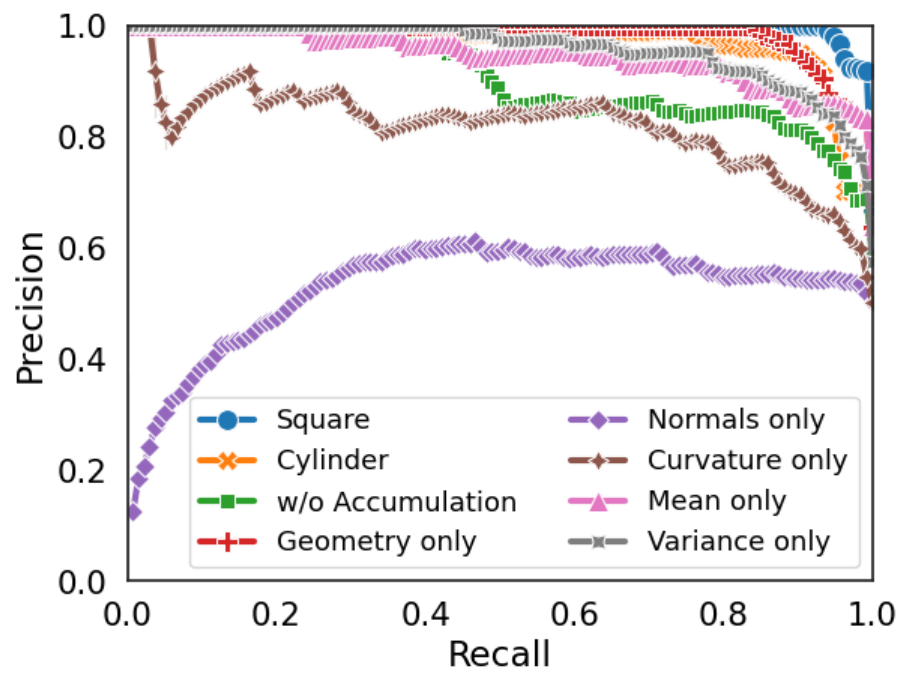}}}
            \label{fig:pr_curve_ablation_study}
        }
    \caption{Precision and recall curves of our method and baselines. From the top left to the bottom right in a zigzag order,} Antarctica dataset, Seaward dataset (including Wiggles bank, North river, Dutch island, Beach pond) and an ablation study.
    \label{fig:pr_curves}
\end{figure*}
In contrast, for the Seaward dataset, due to its shallow depth, the lateral spread of the vehicle-based point cloud was narrower, thus $d$ was set to $10\ m$ for square cropping, as in Eq. \eqref{eqn:square_cropping}. The determination of positive and negative pairs between two sequences was based on the Euclidean distance of the ego-vehicle pose. If the distance was smaller than $0.5\,d$, the pair was considered a positive pair, while distances greater than $2\,d$ were considered negative pairs. For the Seaward dataset, experiments were conducted across four different environments: Wiggles Bank, North River, Dutch Island, and Beach Pond. Moreover, we utilized the Open3D\cite{zhou2018open3d} library to process the point cloud data. Additionally, the auvlib\footnote{https://github.com/nilsbore/auvlib} library was employed for processing the Antarctica dataset.
\subsection{Evaluation}
\inlinesubsubsection{Antarctica}\label{exp:eval_antarctica}
The point cloud and feature map used for loop detection are illustrated in Fig.~\ref{fig:data_examples}. The point cloud and feature map have dimensions of $n \times 3$ and $n \times 1$, respectively. The query feature map is derived from the transformation of the input raw point cloud, while the positive target represents a feature map obtained at the same location as the query feature map but at a different time. In contrast, the negative target refers to a feature map acquired at a different location from the query feature map. The experimental results in the Antarctica dataset are shown in Fig.~\ref{fig:loop_detection_results} and Fig.~\ref{fig:pr_curve_antarctica}. When comparing the results based on Average Precision (AP), which represents the area under the Precision-Recall curve, the AP of our proposed method and \cite{tan2023data} are the highest, confirming that both methods achieve good loop detection performance. \cite{zhang2024shape} shows moderate prediction performance, while \cite{hammond2015automated} demonstrates the lowest performance. From this, we can answer the first research question: unlike \cite{tan2023data} and \cite{zhang2024shape}, which require separate model training or vocabulary creation, our proposed method achieve the best loop detection performance without any additional preprocessing. Furthermore, because \cite{hammond2015automated}, which is based on keypoints for loop detection, does not perform well, the good performance of our proposed method indicates that it works well even in environments where it is difficult to extract appropriate keypoints for loop detection.\\
\indent LiDAR point cloud-based loop detection methods\cite{jiang2023contour,yuan2023std} were excluded from the results as they performed poorly on the Seaward dataset, with loop detection failing in most cases. This poor performance is attributed to the fact that these methods are primarily designed to process dense point clouds obtained from 360-degree LiDAR commonly used in ground vehicle scenarios\cite{jung2023lorcon} or rely on features within the point cloud, such as keypoints or contours. In bathymetry scenarios, where point clouds are extremely sparse, keypoints are almost nonexistent, and there is minimal height variation between points, algorithms that depend on features like keypoints or contours to detect loops are less effective. Instead, as demonstrated in this letter, extracting and comparing point-wise structural feature maps offers significantly better performance.\\
\indent In the case of\cite{kim2018scan}, the method demonstrated a performance level comparable to\cite{zhang2024shape}, a loop detection technique specifically developed for processing sonar data. When loop detection is performed using \cite{kim2018scan}, unlike \cite{jiang2023contour} or\cite{yuan2023std}, it do not rely on keypoints, enabling its functionality even in bathymetric scenarios. In the case of \cite{kim2018scan}, under ground environments, the distribution of points varies significantly based on the heading angle relative to the ego pose, resulting in noticeable differences between scan contexts and thus facilitating loop detection. However, in underwater environments, the point cloud is cropped to a uniform shape of the seafloor to construct the scan context for comparison. This leads to less pronounced differences between sequences, making loop detection more challenging. This limitation is evidenced by the lower accuracy of \cite{kim2018scan} compared to the proposed method, as shown in Fig. \ref{fig:pr_curves}.

\inlinesubsubsection{Seaward}
The performance of each method on the Seaward dataset is shown in Fig.~\ref{fig:loop_detection_results}, Figs.~\ref{fig:pr_curve_wiggles_bank}--\ref{fig:pr_curve_beach_pond}. While there are slight differences in the performance of each algorithm across the four datasets, overall, the proposed method demonstrated the best performance. Therefore, we verify that the proposed method works well in new environments without requiring additional preprocessing, answering the first research question. \cite{tan2023data} and \cite{zhang2024shape} show similar levels of performance, while \cite{hammond2015automated}, as in the previous case with the Antarctica dataset, shows the lowest loop detection performance. The reason for the relatively poor performance of \cite{tan2023data} compared to the Antarctica dataset is attributed to the differences between the Antarctica and Seaward datasets. In the case of the Seaward dataset, data was collected in relatively shallow and flat environments, such as lakes and rivers. As a result, keypoint-based loop detection methods like \cite{hammond2015automated} and \cite{tan2023data} perform poorly on the Seaward dataset. This allows us to address the second research question.\\
\indent Finally, as with the results on the Antarctica dataset, LiDAR-based point cloud methods~\cite{jiang2023contour,yuan2023std} failed to perform loop detection effectively on the Seaward dataset and were therefore excluded from the results. In contrast, \cite{kim2018scan} demonstrated performance comparable to or slightly lower than sonar-based loop detection methods~\cite{tan2023data,zhang2024shape}. These results demonstrate that LiDAR point cloud-based loop detection methods exhibit poor performance not only in deep-sea environments but also across most bathymetric scenarios, including rivers and lakes.

\inlinesubsubsection{Ablation Study}
We conducted an ablation study on the cropping of input data from point clouds. First, as proposed in Sec.~\ref{methods:point_cloud_processing}, we cropped the accumulated point cloud along the ego-vehicle's path into a square shape. This method, applied in \cite{hammond2015automated, zhang2024shape}, which convert point clouds into grid images. Furthermore, referencing \cite{tan2023data}, we performed cylindrical cropping to compare the performance of loop detection. Cylindrical cropping has the advantage that, when passing through a similar location, the point clouds overlap uniformly regardless of the ego-vehicle's orientation. After cropping the point clouds using these two different methods, we compared the results of loop detection. As shown in Fig.~\ref{fig:pr_curve_ablation_study}, it was confirmed that square cropping demonstrated higher loop detection performance than cylindrical cropping. This result can be attributed to the fact that, square cropping includes regions near the corners outside of the circular area, which leads to a larger overlap when the point clouds are misaligned.\\
\indent Furthermore, we evaluated the loop detection performance under various conditions: using the point cloud from a single sequence without accumulation, utilizing only one of the feature maps such as geometry, normals, or curvature, and employing only the mean or variance of the three feature maps. First, in the case of using the point cloud from a single sequence, the number of points in the cloud was fewer than 1,000, making it challenging to accurately detect loops. When using only a single feature map among geometry, normals, or curvature, the geometry feature map yielded the highest performance, indicating that geometry features have a significant impact on overall loop detection accuracy. Lastly, when using only the mean or variance of the feature maps, both showed similar AP performance. This suggests that the mean and variance individually contribute equally to predictive accuracy. However, when combined, they complement each other by mitigating errors in loop predictions in specific cases.
\subsection{Limitation \& Future Works}
In the proposed method, all frames stored in the database are compared with the current frame, resulting in a linear increase in processing time as the number of frames grows. This characteristic implies that for an autonomous underwater vehicle traveling along a long trajectory, the accumulated number of frames will increase, leading to a linear slowdown in the algorithm's speed. To address this limitation, future research will focus on introducing a preliminary candidate extraction process at the database level using vector embedding. This enhancement aims to ensure that the proposed algorithm maintains a consistently fast processing speed even as the trajectory length increases.

\section{Conclusion}
In this letter, we proposed a loop detection algorithm based on the comparison of structural similarity in point clouds obtained through MBES in underwater scenarios. Unlike learning-based methods that require model training or BoW-based methods that require vocabulary creation, our proposed method can accurately detect loops in various environments, such as deep seas, rivers, and lakes, without any additional preprocessing. Moreover, by comparing sequences using the point-wise characteristics of the 3D point cloud without 2D projection, the method could accurately detect loops even in simple environments where it was difficult to extract keypoints from the seafloor. To validate the performance of the proposed method, experiments were conducted using data collected from diverse environments, including the Antarctica and Seaward datasets. The results confirmed that the proposed algorithm is a robust MBES-based underwater loop detection method suitable for various underwater environments.

\bibliographystyle{IEEEtran}
\bibliography{root}
\end{document}